\documentclass{article}

\usepackage{arxiv}

\usepackage[utf8]{inputenc}
\usepackage[T1]{fontenc}
\usepackage[hidelinks]{hyperref}
\usepackage{url}
\usepackage{booktabs}
\usepackage{amsfonts}
\usepackage{nicefrac}
\usepackage{microtype}
\usepackage{graphicx}
\usepackage{multirow}%
\usepackage{amsmath,amssymb,amsfonts}%
\usepackage{amsthm}%
\usepackage{cleveref}
\usepackage[numbers,sort&compress]{natbib}
\usepackage{doi}
\usepackage{orcidlink}
\usepackage{svg}
\usepackage[table]{xcolor} 
\usepackage{float}

\graphicspath{{images/}} 

\title{Enhancing Floor Plan Recognition: A Hybrid Mix-Transformer and U-Net Approach for Precise Wall Segmentation}
\date{\today}
\renewcommand{\shorttitle}{Enhancing Floor Plan Recognition: A Hybrid Mix-Transformer and U-Net Approach}

\hypersetup{
  pdftitle={Enhancing Floor Plan Recognition: A Hybrid Mix-Transformer and U-Net Approach for Precise Wall Segmentation},
  pdfsubject={cs.CV, cs.AI},
  pdfauthor={Dmitriy Parashchuk, Alexey Kaspshitskiy, Yuriy Karyakin},
  pdfkeywords={Floor plan analysis, Semantic segmentation, Vectorization, 3D reconstruction, Hybrid architecture, Mix-Transformer, U-Net},
}

\begin{document}

\author{
  Dmitriy Parashchuk\orcidlink{0009-0003-1933-3569}\thanks{Corresponding author.} \\
  Department of Computer Science \\
  Tyumen State University \\
  Tyumen, Russia \\
  \texttt{parashchuk.dmitriii@gmail.com} \\
  \And
  Alexey Kaspshitskiy\orcidlink{0009-0005-9567-1441} \\
  Department of Computer Science \\
  Tyumen State University \\
  Tyumen, Russia \\
  \texttt{stud0000247033@study.utmn.ru}
  \And
  Yuriy Karyakin\orcidlink{0000-0003-2346-402X} \\
  Department of Computer Science \\
  Tyumen State University \\
  Tyumen, Russia \\
  \texttt{y.e.karyakin@utmn.ru}
}

\maketitle

\begin{abstract}
\noindent
Automatic 3D reconstruction of indoor spaces from 2D floor plans necessitates high-precision semantic segmentation of structural elements, particularly walls. However, existing methods often struggle with detecting thin structures and maintaining geometric precision. To address this, we introduce MitUNet, a hybrid neural network designed to bridge the gap between global semantic context and fine-grained structural details. Our architecture combines a Mix-Transformer encoder with a U-Net decoder enhanced with spatial and channel attention blocks. Optimized with the Tversky loss function, this approach achieves a balance between precision and recall, ensuring accurate boundary recovery. Experiments on the CubiCasa5k dataset and the regional dataset demonstrate MitUNet's superiority in generating structurally correct masks with high boundary accuracy, outperforming standard models. This tool provides a robust foundation for automated 3D reconstruction pipelines. To ensure reproducibility and facilitate future research, the source code and the regional dataset are publicly available at \url{https://github.com/aliasstudio/mitunet} and \url{https://doi.org/10.5281/zenodo.17871079}, respectively.
\end{abstract}


\section{Introduction}

The automated generation of 3D building models from 2D floor plans remains a critical challenge in computer vision. Floor plans, serving as schematic blueprints, encode essential information regarding the spatial structure of indoor environments. However, transforming these 2D representations into three-dimensional models through manual processing involves substantial temporal and financial overhead. The prohibitive manual effort required for such modeling imposes severe scalability constraints, limiting its application in large-scale real estate projects or for individual users seeking to visualize personal spaces. Automation of this workflow offers a pathway to significantly reduce costs, accelerate 3D visualization for real estate marketing, and simplify project planning for private needs.

Traditional architectural software tools, such as Autodesk Revit, AutoCAD, SketchUp, or ArchiCAD, are primarily tailored for manual design and Building Information Modeling (BIM) workflows. These platforms do not inherently support the fully automatic reconstruction of 3D models from raw 2D raster floor plans. Instead, they require professional expertise and significant user intervention to trace and define geometry, rendering them unsuitable for high-throughput automation. Consequently, there is a growing demand for intelligent systems capable of parsing raster schematics autonomously.

This study prioritizes the semantic segmentation of walls as a fundamental prerequisite for constructing structurally coherent 3D room models. Accurate recognition of structural boundaries is critical, as walls determine the topology of the entire building. Segmentation artifacts at this stage, such as discontinuities in wall segments or boundary noise, inevitably compromise the subsequent 3D generation phase, leading to topological inconsistencies or distorted geometry. Our primary objective is to enhance the geometric fidelity of this specific component, establishing a reliable structural basis for vectorization algorithms used in Scan-to-BIM pipelines.

To achieve high-precision recognition, we utilized a custom dataset of 500 floor plans representing distinct regional architectural styles. These plans differ visually and structurally from the samples found in standard open datasets, such as CubiCasa5k \cite{cubicasa5k}, often presenting unique challenges in hatching patterns and layout conventions. Addressing these variations requires a model capable of robust generalization across different drawing domains, forcing us to move beyond standard baseline approaches.

To address the limitations of existing segmentation methods, we propose MitUNet, a hybrid architecture that combines the strengths of two paradigms: a Mix-Transformer encoder derived from SegFormer \cite{segformer2021} to capture global semantic context, and a CNN-based U-Net decoder \cite{unet2015} to precisely recover fine-grained structural details. Furthermore, we support this architecture with a refined optimization strategy utilizing the Tversky loss function. This allows us to explicitly manage the trade-off between recall and precision, ensuring that the model detects thin wall segments while maintaining sharp, accurate boundaries essential for vectorization.

\section{Related Work}

The task of floor plan analysis and understanding has been extensively explored in recent years, with numerous studies proposing pipelines for converting 2D images into structured topological graphs or 3D models. A common thread among these works is the reliance on deep learning to extract geometric primitives, such as rooms, walls, doors, and windows.

Early deep learning approaches primarily utilized standard Convolutional Neural Networks (CNNs). For instance, Zeng et al. \cite{zeng2019deep} proposed a multi-task network leveraging room-boundary-guided attention to simultaneously recognize rooms and boundaries. Their approach relies on a VGG-16 backbone, a classic CNN architecture which, while effective for general feature extraction, often struggles with the long-range dependencies required to distinguish structural walls from decorative lines in complex, cluttered drawings.

More recently, research has shifted towards understanding topological relationships and refining structural segmentation. Huang et al. \cite{muranet2023} introduced MuraNet, utilizing relation attention mechanisms to better understand the relationships between room types. Similarly, Kratochvila et al. \cite{multiunit2024} extended reconstruction capabilities to multi-unit floor plans. In the context of refined segmentation, Yang et al. \cite{exploring2025} recently proposed exploring structural lines specifically for interior floor plan segmentation, demonstrating the importance of geometric priors in this domain. Furthermore, addressing the robustness of segmentation under noise—a common issue in scanned documents—remains a critical area of study, similar to approaches seen in 3D point cloud segmentation under label noise by Zhang et al. \cite{joint2025}.

Despite these advances, pixel-wise segmentation of walls remains challenging. Conventional CNN-based architectures often suffer from limited receptive fields. To mitigate this, attention mechanisms have become pivotal. For example, efficient attention pyramid transformers, as discussed by \cite{eapt2021}, have shown significant promise in general image processing by capturing multi-scale context. In the specific domain of wall detection, Eldosoky et al. \cite{wallnet2025} introduced WallNet, a hierarchical visual attention-based model, to precisely detect terminal points and bulges, highlighting the necessity of hierarchical features for structural fidelity. Additionally, advanced matching techniques, such as the deformable sparse-to-dense feature matching proposed by Zhao et al. \cite{dsd2022}, illustrate the trend towards more adaptive feature extraction methods.

Our work aims to bridge the gap between global context and fine-grained structural details by introducing a hybrid Transformer-CNN architecture. We leverage the hierarchical nature of Mix-Transformers effectively, inspired by the success of attention-based models and approaches that explicitly incorporate geometric priors, to serve as an enhanced segmentation module within broader reconstruction frameworks.

\section{Dataset and Data Preparation}

\subsection{Datasets}
Our study leverages two distinct datasets to ensure generalizability and domain-specific accuracy.

\textbf{CubiCasa5k:} For the pre-training phase, we utilized the CubiCasa5k dataset \cite{cubicasa5k}, a large-scale collection of 5,000 diverse floor plan images. This dataset is characterized by a high degree of variability in drawing styles, image quality, and clutter levels, effectively acting as a noisy source of structural data. Training on such diverse data allows the model's backbone to learn robust, invariant geometric features and understand the general topology of indoor spaces, preventing overfitting to a single drawing convention.

\textbf{Regional Dataset:} To evaluate the model's performance on real-world data distinct from the training distribution, we utilized the \textbf{Floor Plan CIS} dataset \cite{floorplancis2025}. This dataset, curated by the authors, consists of 500 floor plans collected from publicly available real estate listings within the Russian Federation and CIS region. To ensure reproducibility and facilitate future research on domain adaptation, it has been publicly released via Zenodo (DOI: \url{https://doi.org/10.5281/zenodo.17871079}).

Visual analysis of this dataset reveals significant domain shifts compared to the standard CubiCasa5k benchmark:
\begin{itemize}
    \item \textbf{Texture-based Material Encoding:} Unlike the uniform wall fills in Western datasets, these plans strictly differentiate construction materials. Load-bearing structures are often depicted as solid black fills, while partition walls feature complex internal hatching (e.g., diagonal strokes). Standard segmentation models frequently misclassify these textured regions as background noise.
    \item \textbf{Complex Geometry and Topology:} The layouts frequently feature non-Manhattan geometries (curved or angled outer walls) and varied structural topologies which challenge the geometric priors of models trained solely on rectangular boxes.
    \item \textbf{High-Density Clutter:} The plans contain dense semantic noise, including furniture outlines, dimension lines, and text overlays directly on the floor space.
\end{itemize}
All images were manually annotated to generate precise polygonal masks, strictly isolating the structural wall geometry from decorative elements.

\begin{figure}[htbp] 
    \centering
    \includegraphics[width=\textwidth]{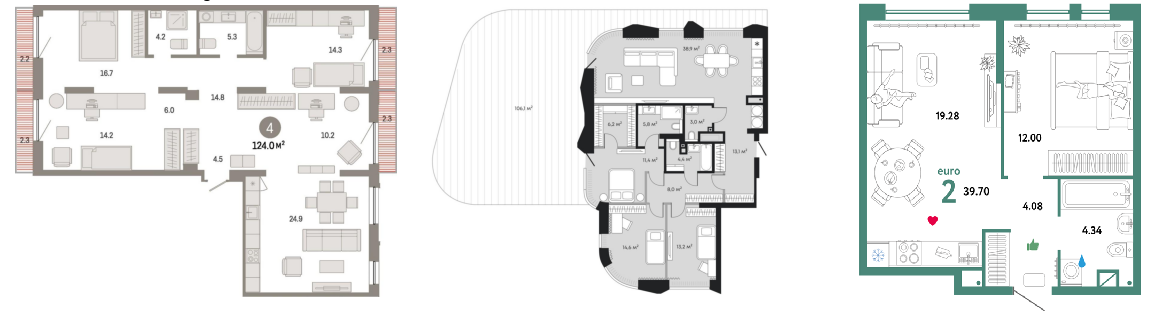} 
    \caption{Representative samples from our collected Regional Dataset demonstrating key segmentation challenges: complex wall hatching patterns (differentiating partitions from load-bearing walls), non-Manhattan geometry, and dense semantic clutter (furniture, text, and dimension lines).}
    \label{fig:dataset_samples}
\end{figure}

\subsection{Data Preprocessing}
To optimize computational efficiency and ensure training stability on a single GPU workstation, we standardized the input resolution to $512 \times 512$ pixels. This resolution was empirically determined to provide sufficient spatial granularity for resolving thin wall structures and hatching patterns while maintaining a batch size of 4, which is critical for effective Batch Normalization statistics given the memory constraints of the model.

A critical aspect of our pipeline is the \textit{annotation refinement} procedure. We observed that ground truth annotations in public datasets like CubiCasa5k often exhibit inconsistencies, such as overlapping polygons for walls and openings (doors, windows). Such overlaps introduce ambiguity during training, as the model receives conflicting signals for pixels that belong to both \textit{wall} and \textit{door} classes. To resolve this and ensure the model learns to segment only the solid structural components, we implemented a robust subtraction procedure:
\begin{enumerate}
    \item We generate separate binary masks for doors and windows.
    \item To account for potential annotation inaccuracies and ensure complete removal of openings from the wall mask, we structurally dilate these opening masks by a fixed margin (approx. 30 pixels).
    \item These dilated opening masks are subtracted from the wall mask, effectively carving out precise holes where doors and windows are located. This ensures that the resulting ground truth strictly represents the solid wall geometry.
    \item Finally, we apply a morphological closing operation (using a $5 \times 5$ kernel) to remove small artifacts and ensure wall continuity.
\end{enumerate}

\section{Methodology}

\subsection{Architecture: MitUNet}
\begin{figure}[t]
    \centering
    \includegraphics[width=\textwidth]{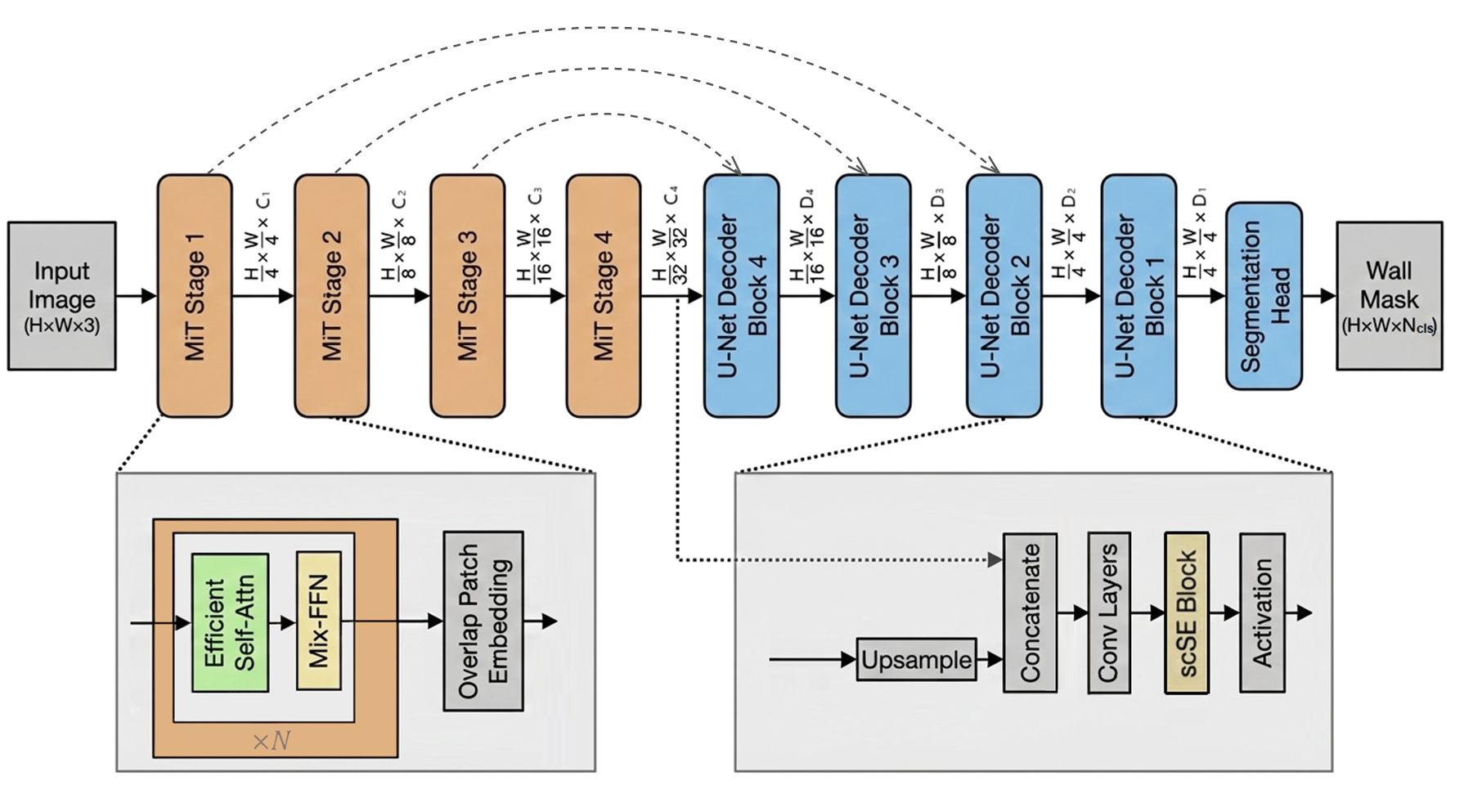}
    \caption{Architecture of the proposed MitUNet. The model combines a hierarchical Mix-Transformer encoder ($C_i=\{64, 128, 320, 512\}$) with a U-Net decoder ($D_i=\{32, 64, 128, 256\}$). The inset (bottom right) details the decoding stage, where scSE attention is applied after convolution to refine boundary features.}
    \label{fig:architecture}
\end{figure}

The semantic segmentation of floor plans presents a dual challenge: the model must possess a large effective receptive field to comprehend the global structural context, while simultaneously maintaining high spatial resolution to precisely delineate thin elements like walls. To address this, we propose MitUNet, a hybrid architecture illustrated in Fig. \ref{fig:architecture}, which combines a Transformer-based encoder for global context with a CNN-based attention decoder for fine-grained structural details.

\textbf{Hierarchical Mix-Transformer Encoder:} 
As the feature extractor, we utilize the MiT-b4 backbone from the SegFormer framework \cite{segformer2021}, pre-trained on ImageNet \cite{imagenet}. Unlike standard Vision Transformers (ViT) that generate single-scale feature maps, the Mix-Transformer (MiT) adopts a hierarchical design similar to CNNs. It produces multi-scale features at resolutions of $\{1/4, 1/8, 1/16, 1/32\}$ relative to the input image, with corresponding channel dimensions $C_i=\{64, 128, 320, 512\}$. We selected the \texttt{b4} variant because it offers an optimal trade-off between computational efficiency and capacity. Crucially, the encoder employs an overlapping patch merging mechanism. Unlike non-overlapping patches in standard ViTs which can introduce grid artifacts, this design preserves local continuity—a critical factor for tracing continuous wall geometries without artificial discontinuities.

\textbf{High-Resolution Attention Decoder:} 
A critical limitation of the standard SegFormer architecture is its lightweight MLP decoder, which aggregates features at a coarse resolution ($\frac{1}{4}$) and relies on bilinear upsampling to restore the final mask. For thin wall segmentation, this approach inherently suffers from aliasing and the loss of high-frequency edge information.
To overcome this, we discard the MLP decoder in favor of a full U-Net \cite{unet2015} reconstruction path. This decoder progressively upsamples the feature maps, fusing them with high-resolution features from the encoder via skip connections. To balance representational efficiency, we set the decoder channel dimensions to $D_i=\{256, 128, 64, 32\}$ from the deepest to the shallowest block. This dense reconstruction path is essential to explicitly recover the spatial details lost during downsampling, which simple interpolation cannot restore.

Furthermore, we enhance each decoding stage with scSE (Spatial and Channel Squeeze \& Excitation) blocks \cite{scse2018}, integrated after the convolutional layers (see inset in Fig. \ref{fig:architecture}). The scSE module adaptively recalibrates the feature maps:
\begin{itemize}
    \item \textbf{Spatial Squeeze:} Highlights pixels relevant to wall locations, suppressing background noise.
    \item \textbf{Channel Squeeze:} Emphasizes feature maps that carry the most relevant semantic information (e.g., texture vs. shape).
\end{itemize}
This combination ensures that the global context captured by the Transformer is effectively translated into pixel-perfect local boundaries.

\subsection{Loss Function Strategy}
The choice of loss function is pivotal for segmenting thin, imbalanced classes like walls. To explicitly control the trade-off between Precision and Recall, we evaluated several standard approaches. Let $N$ be the total number of pixels, $p_{i} \in [0,1]$ denote the predicted probability of pixel $i$ belonging to the wall class, and $g_{i} \in \{0,1\}$ be the corresponding ground truth label.

\textbf{Dice Loss:} Proposed by Milletari et al. \cite{milletari2016vnet}, this loss directly optimizes the F1 score (Dice coefficient) and inherently addresses class imbalance without requiring weight assignment.
\begin{equation}
    L_{Dice} = 1 - \frac{2 \sum_{i=1}^{N} p_{i} g_{i} + \epsilon}{\sum_{i=1}^{N} p_{i} + \sum_{i=1}^{N} g_{i} + \epsilon}
\end{equation}
where $\epsilon$ is a small smoothing term added for numerical stability to prevent division by zero. However, solely maximizing overlap can sometimes encourage the model to output dilated masks, reducing boundary sharpness.

\textbf{Focal Loss:} Introduced by Lin et al. \cite{lin2017focal}, this loss addresses the extreme class imbalance by reducing the weight of easily classified examples. It adds a modulating factor $(1 - p_t)^\gamma$ to the standard cross-entropy:
\begin{equation}
    L_{Focal} = - \frac{1}{N} \sum_{i=1}^{N} \alpha_t (1 - p_{t,i})^\gamma \log(p_{t,i})
\end{equation}
where $p_{t,i}$ reflects the probability of the ground truth class ($p_i$ if $g_i=1$, and $1-p_i$ otherwise), $\gamma$ is the focusing parameter, and $\alpha_t$ is a balancing factor. This formulation prevents the vast number of easy negatives (background) from overwhelming the detector during training.

\textbf{Lovasz-Softmax Loss:} Proposed by Berman et al. \cite{berman2018lovasz}, this method serves as a convex surrogate to directly optimize the Intersection-over-Union (IoU). It minimizes the Lovasz extension of the Jaccard loss:
\begin{equation}
    L_{Lovasz} = \frac{1}{|C|} \sum_{c \in C} \overline{\Delta_{J_c}}(\boldsymbol{m}(c))
\end{equation}
where $|C|$ is the number of classes, $\overline{\Delta_{J_c}}$ denotes the Lovasz extension of the set function $\Delta_{J_c}$ (Jaccard loss), and $\boldsymbol{m}(c)$ represents the vector of pixel errors for class $c$, defined as:
\begin{equation}
    m_i(c) = 
    \begin{cases} 
    1 - p_i(c) & \text{if } g_i = c \\
    p_i(c) & \text{otherwise}
    \end{cases}
\end{equation}
This approach allows for direct optimization of the global metric, but lacks the explicit control over boundary precision provided by the Tversky parameters.

\textbf{Tversky Loss:} To strictly control the balance between boundary precision and structural completeness, we employ the Tversky loss \cite{tversky2017}. Based on the Tversky index, this loss generalizes the Dice coefficient by introducing asymmetric penalties for false positives (FP) and false negatives (FN):
\begin{equation}
    L_{Tversky}(\alpha, \beta) = 1 - \frac{\sum_{i=1}^{N} p_{i} g_{i} + \epsilon}{\sum_{i=1}^{N} p_{i} g_{i} + \alpha \sum_{i=1}^{N} p_{i} (1-g_{i}) + \beta \sum_{i=1}^{N} (1-p_{i}) g_{i} + \epsilon}
\end{equation}
Here, $\alpha$ and $\beta$ are hyperparameters that control the trade-off between FP and FN, and $\epsilon$ is a smoothing factor. By setting $\alpha > \beta$, we penalize false positives more heavily, effectively suppressing noise and "thickened" boundaries.

In our experimental framework, we conduct a comparative analysis of these four objective functions to identify the optimal strategy for thin-wall segmentation. While Dice and Lovasz-Softmax focus on maximizing global overlap, they lack the mechanism to penalize specific error types explicitly. We hypothesize that the Tversky loss, with its tunable asymmetric penalties ($\alpha > \beta$), offers a distinct advantage by actively suppressing false positive noise along boundaries—a critical requirement for generating clean vectorization-ready masks.

\section{Experiments}

\subsection{Experimental Setup}
\label{sec:experimental_setup}

All experiments were conducted on a workstation equipped with a NVIDIA RTX 4060 Ti (16 GB VRAM). The training pipeline was implemented using the PyTorch framework \cite{pytorch2019}, utilizing the Albumentations library \cite{albumentations2020} for data preprocessing.

To ensure robust evaluation, we aggregated available data into a single pool and performed a randomized stratified split. We utilized an 80/20 ratio for training and validation, respectively, governed by a fixed random seed (42). Input images were standardized to $512 \times 512$ pixels and normalized using ImageNet \cite{imagenet} mean ($\mu=[0.485, 0.456, 0.406]$) and standard deviation ($\sigma=[0.229, 0.224, 0.225]$).

\textbf{Augmentation Strategy:} To improve robustness to scan quality variations and geometric distortions, we applied a comprehensive dynamic augmentation pipeline:
\begin{itemize}
    \item \textbf{Geometric Transformations:} Random affine scaling ($\in [0.9, 1.1]$), rotation ($\in [-15^\circ, 15^\circ]$), and translation ($p=0.7$). We also included elastic transformations and grid distortions ($p=0.2$) to mimic paper warping common in scanned documents.
    \item \textbf{Photometric Transformations:} Random brightness/contrast adjustments ($p=0.5$) and CLAHE ($p=0.2$) were applied to ensure invariance to lighting conditions. Gaussian and ISO noise were introduced to simulate sensor artifacts.
\end{itemize}

\textbf{Training Protocol:}
We adopted a standardized protocol for all model variations:
\begin{itemize}
    \item \textbf{Optimization:} Adam optimizer \cite{adam2014} with an initial learning rate determined empirically.
    \item \textbf{Strategy:} We used a two-stage transfer learning approach. First, models were pre-trained on the diverse CubiCasa5k dataset \cite{cubicasa5k} to learn general geometric features. Subsequently, they were fine-tuned on the Regional Dataset to adapt to specific architectural styles.
    \item \textbf{Parameters:} Training ran for 30 epochs per stage with a batch size of 4. A \texttt{ReduceLROnPlateau} scheduler (factor=0.5, patience=3) monitored validation IoU.
\end{itemize}
Each experiment was repeated three times to mitigate initialization effects; we report average metrics for the best validation IoU.

\subsection{Comparative Analysis}
We first evaluated the performance of our proposed MitUNet architecture against several state-of-the-art segmentation models, including UNet++ \cite{unetplusplus2018} (with ResNet50 backbone), DeepLabV3+ \cite{deeplab2018}, SegFormer \cite{segformer2021}, and UPerNet \cite{upernet2018}. We also tested various loss functions to establish a baseline.

\begin{table}[htbp]
\centering
\caption{Quantitative comparison on the Regional Dataset. Results are grouped by architecture type and sorted by mIoU. The inclusion of Boundary IoU (B-IoU) explicitly highlights MitUNet's superiority in preserving sharp structural edges.}
\label{tab:sota_comparison}
\scriptsize 
\setlength{\tabcolsep}{15pt}
\renewcommand{\arraystretch}{1.1}

\begin{tabular}{@{}llccccccc@{}}
\toprule
\textbf{Model} & \textbf{Encoder} & \textbf{Loss} & \textbf{Recall} & \textbf{Precision} & \textbf{Accuracy} & \textbf{mIoU} & \textbf{B-IoU} & \textbf{VRAM} \\ 
& & & \textbf{(\%)} & \textbf{(\%)} & \textbf{(\%)} & \textbf{(\%)} & \textbf{(\%)} & \textbf{(MiB)} \\
\midrule
\multicolumn{9}{c}{\cellcolor{gray!10}\textit{\textbf{Transformer-based Architectures}}} \\
\midrule
\multirow{4}{*}{\textbf{MitUNet}} & \multirow{4}{*}{mit\_b4} 
 & Tversky & 92.35 & \textbf{94.82} & \textbf{98.86} & \textbf{87.91} & \textbf{85.01} & 1751 \\
 & & Dice & 93.28 & 93.51 & 98.81 & 87.61 & 84.89 & 1751 \\
 & & Lovasz & 93.56 & 92.89 & 98.78 & 87.31 & 84.48 & 1751 \\
 & & Focal & 92.74 & 93.29 & 98.75 & 86.94 & 84.15 & 1751 \\
\addlinespace
\multirow{4}{*}{UPerNet} & \multirow{4}{*}{mit\_b4}
 & Dice & 92.97 & 92.75 & 98.71 & 86.67 & 82.91 & 2211 \\
 & & Lovasz & 92.21 & 93.09 & 98.68 & 86.31 & 83.03 & 2219 \\
 & & Tversky & 91.55 & 93.62 & 98.68 & 86.18 & 82.36 & 2219 \\
 & & Focal & 93.06 & 91.65 & 98.61 & 85.79 & 82.42 & 2219 \\
\addlinespace
\multirow{4}{*}{SegFormer} & \multirow{4}{*}{mit\_b4}
 & Lovasz & 93.88 & 91.75 & 98.69 & 86.57 & 83.26 & 1270 \\
 & & Dice & 92.57 & 92.44 & 98.65 & 86.05 & 82.47 & 1270 \\
 & & Focal & 92.58 & 92.00 & 98.61 & 85.68 & 82.21 & 1270 \\
 & & Tversky & 90.85 & 93.44 & 98.60 & 85.41 & 81.31 & 1270 \\
\midrule
\multicolumn{9}{c}{\cellcolor{gray!10}\textit{\textbf{CNN-based Architectures}}} \\
\midrule
\multirow{4}{*}{UNet++} & \multirow{4}{*}{resnet50}
 & Lovasz & 93.74 & 92.65 & 98.77 & 87.25 & 84.68 & 3311 \\
 & & Dice & \textbf{94.09} & 92.26 & 98.76 & 87.21 & 84.62 & 3311 \\
 & & Focal & 92.41 & 93.41 & 98.73 & 86.76 & 83.77 & 3311 \\
 & & Tversky & 92.38 & 93.44 & 98.73 & 86.75 & 84.00 & 3311 \\
\addlinespace
\multirow{4}{*}{UNet scSE} & \multirow{4}{*}{resnet50}
 & Dice & 93.63 & 92.34 & 98.73 & 86.87 & 84.25 & 1503 \\
 & & Lovasz & 92.79 & 92.85 & 98.71 & 86.60 & 83.90 & 1503 \\
 & & Tversky & 91.68 & 93.90 & 98.72 & 86.53 & 83.87 & 1503 \\
 & & Focal & 91.27 & 93.11 & 98.61 & 85.50 & 82.47 & 1503 \\
\addlinespace
\multirow{4}{*}{DeepLabV3+} & \multirow{4}{*}{resnet50}
 & Lovasz & 91.42 & 92.56 & 98.57 & 85.16 & 81.59 & \textbf{947} \\
 & & Dice & 90.47 & 92.61 & 98.49 & 84.38 & 80.25 & \textbf{947} \\
 & & Focal & 91.41 & 91.40 & 98.45 & 84.17 & 80.65 & \textbf{947} \\
 & & Tversky & 89.80 & 92.47 & 98.42 & 83.69 & 79.61 & \textbf{947} \\
\bottomrule
\multicolumn{9}{l}{\scriptsize * Tversky params: $\alpha=0.6, \beta=0.4$. "Tversky" in table denotes this config.}
\end{tabular}
\end{table}

As shown in Table \ref{tab:sota_comparison}, MitUNet with Tversky loss achieved the highest mIoU (87.91\%) and Precision (94.82\%). While purely convolutional models like UNet++ performed competitively (87.25\% mIoU), they incurred nearly double the memory cost (3311 MiB vs. 1751 MiB for MitUNet). 

\textbf{Boundary Quality Analysis:} To rigorously address the challenge of precise edge alignment and explicitly validate the necessity of our high-resolution decoder, we adopted the Boundary IoU (B-IoU) metric \cite{cheng2021boundary}. Conventional mIoU is often dominated by internal pixels, masking boundary dilation artifacts. Our quantitative evaluation reveals a critical interplay between architectural capacity and loss formulation. For instance, when trained with the asymmetric Tversky loss ($\alpha=0.6$), the baseline SegFormer exhibited a severe degradation in boundary quality, achieving the lowest B-IoU among its variants (81.31\%). This occurs because SegFormer's lightweight MLP decoder relies on bilinear upsampling and lacks the spatial resolution to generate thin boundaries, leading to fragmented masks when strictly penalized for false positives. Conversely, MitUNet explicitly utilizes a high-resolution U-Net decoder with scSE blocks to recover these spatial details. When guided by the same Tversky loss, MitUNet achieved the highest overall B-IoU of 85.01\%. This explicitly demonstrates that the U-Net decoder is not redundant, but strictly essential to physically realize the sharp, vectorization-ready boundaries demanded by the asymmetric loss constraint.

\begin{figure}[htbp]
    \centering
    \includegraphics[width=\textwidth]{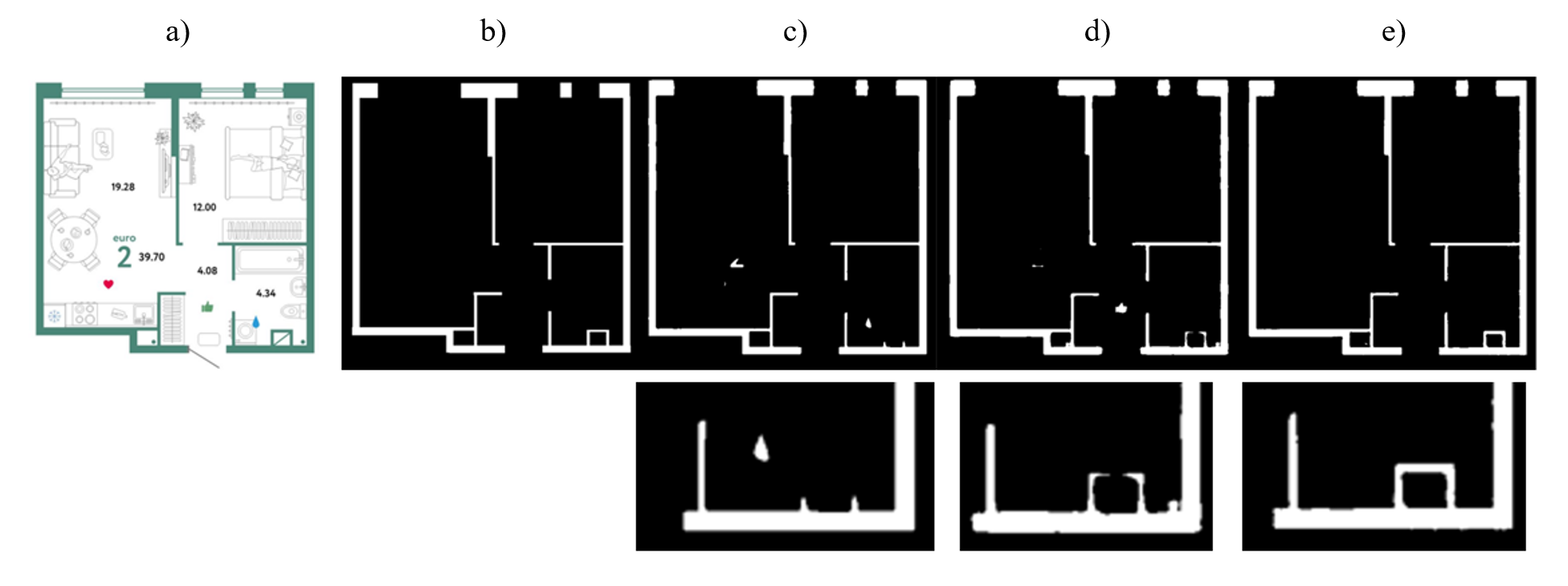}
    \caption{Qualitative comparison of segmentation results on the Regional Dataset. \textbf{(a)} Original input; \textbf{(b)} Ground Truth; \textbf{(c)} UNet (scSE); \textbf{(d)} SegFormer; \textbf{(e)} MitUNet (Ours). The bottom row displays zoomed-in details corresponding to models c, d, and e. Note that UNet introduces noise artifacts (center crop), and SegFormer suffers from dilated or blurred boundaries, whereas MitUNet successfully suppresses noise while maintaining sharp structural edges.}
    \label{fig:arch_comparison}
\end{figure}

\subsection{Tversky Loss Hyperparameter Tuning}
Having established the architectural superiority of MitUNet, we tested our hypothesis that the Tversky loss could provide better control over the Precision/Recall trade-off than standard losses. We conducted an ablation study varying the $\alpha$ (penalty for False Positives) and $\beta$ (penalty for False Negatives) parameters.

\begin{figure}[htbp]
    \centering
    \includegraphics[width=\textwidth]{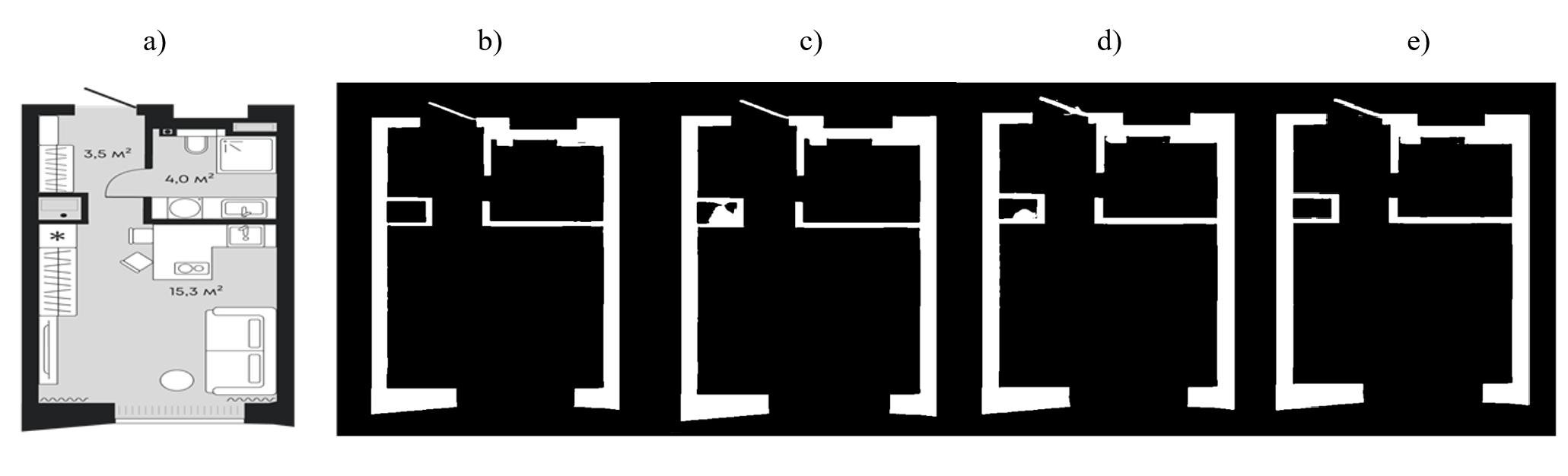}
    \caption{Visual comparison of loss functions during the ablation phase. \textbf{(a)} Original input; \textbf{(b)} Tversky loss effectively suppresses false positives, leaving the ventilation shaft (center-left) correctly hollow; \textbf{(c)} Dice Loss hallucinates a solid block inside the shaft and dilates wall thickness; \textbf{(d)} Focal Loss exhibits similar internal noise artifacts; \textbf{(e)} Lovasz-Softmax preserves basic structure but introduces minor artifacts. This comparison highlights the robust "cleaning" effect of the asymmetric Tversky loss on semantic clutter.}
    \label{fig:loss_ablation}
\end{figure}

The results in Table \ref{tab:tversky_ablation} confirm that increasing $\alpha$ significantly increases Precision at the cost of Recall. The configuration with $\alpha=0.6, \beta=0.4$ achieved the highest mIoU (87.91\%) and B-IoU (85.01\%), maintaining a strong balance between Recall (92.35\%) and Precision (94.82\%). By explicitly penalizing false positives ($\alpha=0.6$), the asymmetric Tversky formulation effectively filters out semantic noise (Fig. \ref{fig:loss_ablation}b), leaving ventilation shafts correctly hollow and preventing text overlays from being registered as physical structures. Therefore, we selected $\alpha=0.6, \beta=0.4$ as the optimal configuration for the subsequent fine-tuning stage.

\begin{table}[htbp]
\centering
\caption{Ablation Study of Tversky Loss Parameters on MitUNet. Increasing $\alpha$ improves Precision significantly but degrades B-IoU if the boundaries become overly thinned.}
\label{tab:tversky_ablation}
\scriptsize 
\setlength{\tabcolsep}{15pt} 
\renewcommand{\arraystretch}{1.1}

\begin{tabular}{llcccccc}
\toprule
\textbf{Model} & \textbf{Encoder} & \textbf{Loss} & \textbf{Recall} & \textbf{Precision} & \textbf{Accuracy} & \textbf{mIoU} & \textbf{B-IoU} \\ 
& & \textbf{($\alpha / \beta$)} & \textbf{(\%)} & \textbf{(\%)} & \textbf{(\%)} & \textbf{(\%)} & \textbf{(\%)} \\
\midrule
\rowcolor{gray!15} MitUNet & mit\_b4 & Tversky (0.6 / 0.4) & \textbf{92.35} & 94.82 & \textbf{98.86} & \textbf{87.91} & \textbf{85.01} \\
MitUNet & mit\_b4 & Tversky (0.7 / 0.3) & 89.20 & 96.25 & 98.71 & 86.20 & 82.75 \\
MitUNet & mit\_b4 & Tversky (0.8 / 0.2) & 87.13 & 97.14 & 98.61 & 84.95 & 81.38 \\
MitUNet & mit\_b4 & Tversky (0.9 / 0.1) & 81.41 & \textbf{98.23} & 98.20 & 80.23 & 75.40 \\
\bottomrule
\end{tabular}
\end{table}

\subsection{Fine-tuning Strategy and Final Results}
Finally, we evaluated the impact of our transfer learning strategy. We hypothesized that exposing the model to the large-scale diversity of the CubiCasa5k dataset would enable it to learn robust geometric priors before domain adaptation. We initialized the fine-tuning process using the optimal MitUNet checkpoint pre-trained on the CubiCasa5k dataset. We then fine-tuned this model on our target Regional Dataset using a reduced learning rate ($1e-5$).

\begin{table}[htbp]
\centering
\caption{Results of Fine-Tuning (initialized with CubiCasa5k pre-trained weights) on the Regional Dataset.}
\label{tab:finetune_results}
\scriptsize 
\setlength{\tabcolsep}{15pt} 
\renewcommand{\arraystretch}{1.1}

\begin{tabular}{llcccccc}
\toprule
\textbf{Model} & \textbf{Encoder} & \textbf{Loss} & \textbf{Recall} & \textbf{Precision} & \textbf{Accuracy} & \textbf{mIoU} & \textbf{B-IoU} \\ 
\textit{(Pre-trained)} & & \textbf{($\alpha / \beta$)} & \textbf{(\%)} & \textbf{(\%)} & \textbf{(\%)} & \textbf{(\%)} & \textbf{(\%)} \\
\midrule
MitUNet & mit\_b4 & Lovasz & \textbf{94.48} & 93.60 & \textbf{98.93} & \textbf{88.75} & 85.93 \\
MitUNet & mit\_b4 & Dice & 94.42 & 93.58 & 98.91 & 88.67 & \textbf{85.98} \\
\rowcolor{gray!15} MitUNet & mit\_b4 & Tversky (0.6 / 0.4) & 92.87 & 94.99 & 98.92 & 88.53 & 85.70 \\
MitUNet & mit\_b4 & Focal & 93.97 & 93.60 & 98.88 & 88.29 & 85.21 \\
MitUNet & mit\_b4 & Tversky (0.7 / 0.3) & 91.29 & 96.00 & 98.87 & 87.95 & 84.93 \\
MitUNet & mit\_b4 & Tversky (0.8 / 0.2) & 88.41 & 97.30 & 98.74 & 86.30 & 82.99 \\
MitUNet & mit\_b4 & Tversky (0.9 / 0.1) & 83.74 & \textbf{98.48} & 98.42 & 82.67 & 78.86 \\
\bottomrule
\end{tabular}
\end{table}

The fine-tuning results (Table \ref{tab:finetune_results}) demonstrate a substantial improvement across all metrics compared to training from scratch. Although the Lovasz and Dice losses achieved marginally higher mIoU and B-IoU scores, the model trained with Tversky loss ($\alpha=0.6, \beta=0.4$) demonstrated a significantly superior Precision of 94.99\% (compared to $\sim 93.6\%$ for Dice and Lovasz). This trade-off is deliberate: while symmetric losses maximize overlap by sometimes dilating boundaries, the asymmetric Tversky loss effectively suppresses false positive noise, providing the geometric definition required for high-quality vectorization.

\begin{figure}[htbp]
    \centering
    \includegraphics[width=\textwidth]{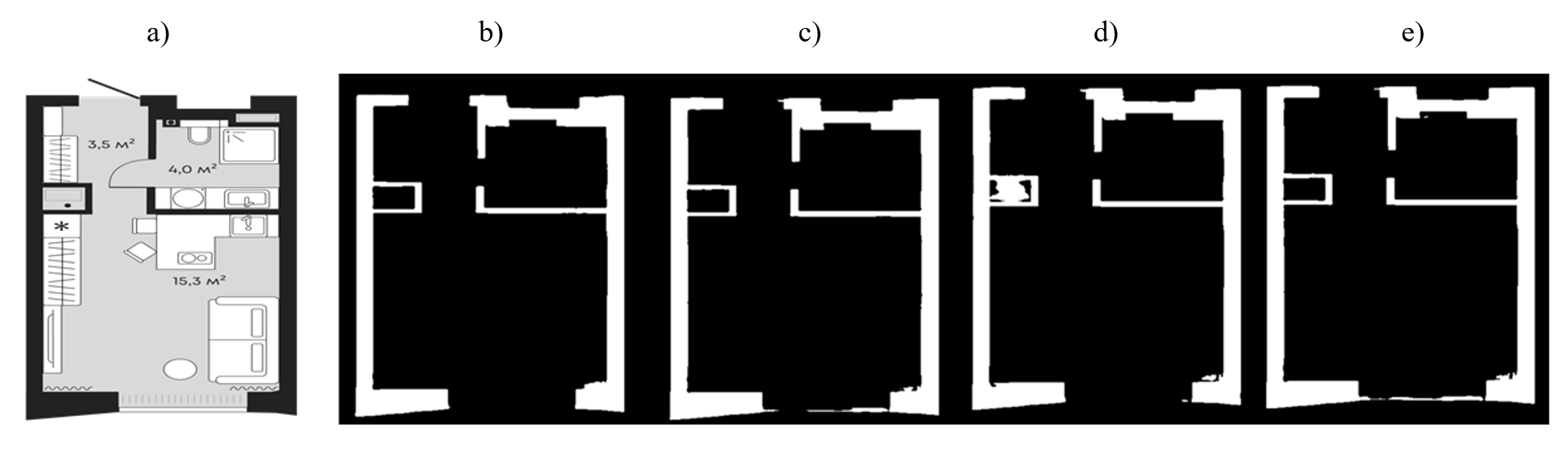}
    \caption{Qualitative comparison of fine-tuned models. \textbf{(a)} Original input; \textbf{(b)} MitUNet trained with Tversky $\alpha=0.6, \beta=0.4$ demonstrates the optimal balance of connectivity and sharpness; \textbf{(c)} Dice Loss; \textbf{(d)} Focal Loss; \textbf{(e)} Lovasz-Softmax. Comparison reveals that our method (b) minimizes "staircase" artifacts along edges compared to standard losses.}
    \label{fig:finetune_comparison}
\end{figure}

\textbf{Impact on Vectorization Quality:} Beyond standard pixel-wise metrics, the choice of the asymmetric Tversky loss provided a critical qualitative advantage. We observed that standard symmetric losses (like Dice) tend to produce dilated boundaries to maximize overlap, resulting in staircase artifacts along wall edges. By penalizing false positives more heavily ($\alpha=0.6$), our optimized MitUNet produces thinner and sharper high-confidence masks (visualized in Fig. \ref{fig:finetune_comparison}b). This geometric crispness significantly reduces noise for downstream vectorization algorithms, minimizing the need for aggressive post-processing smoothing which often distorts corner geometry.

\section{Conclusion}

In this study, we introduced MitUNet, a hybrid segmentation architecture designed to address the specific challenges of floor plan analysis. By combining a hierarchical Mix-Transformer encoder with a fine-grained U-Net decoder, our approach seeks to reconcile global semantic understanding with the pixel-level accuracy required for structural walls.

Experimental evaluation on both the public CubiCasa5k benchmark and our Regional Dataset yielded several key insights. First, the MitUNet architecture demonstrated superior boundary precision compared to baseline CNN models (such as UNet++) and pure Transformer approaches, particularly in resolving thin wall geometries. Second, we confirmed the importance of the loss function configuration for vectorization-oriented tasks. The asymmetric Tversky loss ($\alpha=0.6, \beta=0.4$) provided an effective mechanism to suppress false positive noise, resulting in cleaner segmentation masks. Finally, our two-stage transfer learning strategy, comprising pre-training on diverse data and fine-tuning on domain-specific samples, proved essential for adapting to complex regional hatching patterns, achieving high performance metrics (Recall $>92\%$, Precision $\approx 95\%$) on the target dataset.

We believe that MitUNet offers a promising foundation for automated Scan-to-BIM pipelines, potentially reducing the manual effort involved in 3D modeling. Future work will focus on extending this module into an end-to-end vectorization framework capable of directly generating topological graphs from raster images.

\section*{Acknowledgments}
This study was supported by the Ministry of Science and Higher Education of the Russian Federation within the framework of a State assignment (FEWZ-2024-0052).

\bibliographystyle{unsrtnat}
\bibliography{references} 

\end{document}